%% file: main.tex
\documentclass[runningheads]{llncs}

\usepackage{eccv}

\usepackage{eccvabbrv}

\usepackage{graphicx}
\usepackage{booktabs}

\usepackage[accsupp]{axessibility}  

\usepackage{hyperref}
\usepackage{multirow}
\usepackage{makecell}
\usepackage{wrapfig}
\usepackage{pifont}

\usepackage{orcidlink}

\begin{document}

\title{Sen-Cap: Sensor-Flexible and Noise-Resilient Human Motion Capture via LiDAR-Camera Integration}

\titlerunning{Sen-Cap}
\newcommand{\equalcontrib}{\ensuremath{^{\dagger}}}


\author{Aoru Xue\inst{1}\equalcontrib \and
Yujing Sun\inst{2}\equalcontrib \and
Yiming Ren\inst{1, 2} \and
Kwok-Yan Lam\inst{2} \and
Mao Ye\inst{3} \and
Yuexin Ma\inst{1}
\thanks{Corresponding author. }}

\authorrunning{Xue and Sun, A. et al.}

\institute{$^1$ShanghaiTech University, China, $^2$ Digital Trust Centre, Nanyang Technological University, Singapore, $^3$ EABOT.AI, China\\
\email{\{xuear2024, mayuexin\}@shanghaitech.edu.cn}}

\makeatletter

\maketitle
\begingroup
\renewcommand{\thefootnote}{\ensuremath{\dagger}}
\footnotetext{Equal contribution.}
\endgroup

\input{sec/0_abstract}    
\input{sec/1_intro}
\input{sec/2_related_work}
\input{sec/3_method}
\input{sec/4_Experiment}
\input{sec/6_Conclusion}

\section*{Acknowledgements }

\noindent This work is supported by the National Natural Science Foundation of China (Project Number 62595774), MoE Key Laboratory of Intelligent Perception and Human-Machine Collaboration (KLIP-HuMaCo), HPC Platform of ShanghaiTech University.

\noindent This research is supported by the National Research Foundation, Singapore and Infocomm Media Development Authority under its Trust Tech Funding Initiative. Any opinions, findings and conclusions or recommendations expressed in this material are those of the author(s) and do not reflect the views of National Research Foundation, Singapore and Infocomm Media Development Authority.

\bibliographystyle{splncs04}
\bibliography{main}
\end{document}

%% file: sec/0_abstract.tex
\begin{abstract}
We propose \textbf{Sen-Cap}, a \underline{\textbf{Se}}nsor-Flexible and \underline{\textbf{N}}oise-Resilient 3D human motion \underline{\textbf{Cap}}ture framework that integrates multi-modal data from LiDAR and camera. While multi-modal sensors provide richer information than single-modal sensors, existing approaches still suffer from two core challenges. First, multi-modal alignment/matching across arbitrarily deployed sensors is typically handled by explicit calibration, which propagates errors under changing viewpoints and in turn constrains deployment to fixed, highly overlapped layouts. Second, prior methods degrade under severe noise or partial sensor failures, which are common in real-world environments. To address these challenges, \textbf{Sen-Cap} introduces a Unified Across-Sensor Motion Estimator that reconstructs local pose and shape in a human-centric space without calibrations between sensors, supporting a flexible number of sensors, as well as a Noise-Resistant Trajectory Tracker that maintains robustness under severe point cloud noise through iterative refinement. These sensor-flexible and noise-resilient features make Sen-Cap more practical in real-world deployment. Notably, operating in real time, \textbf{Sen-Cap} achieves state-of-the-art performance on major metrics on Human-M3 and FreeMotion, as well as strong cross-domain performance on LiDARHuman26M and RELI11D. This combination of flexibility and robustness opens new opportunities for motion capture in real-world scenarios, e.g. sports analytics, field robotics, and large-scale immersive environments.
\end{abstract}

%% file: sec/1_intro.tex
\section{Introduction}
Accurate and robust global 3D human motion capture~\cite{EventCap_CVPR2020, xu2022vitpose, jiang2023rtmpose, kaufmann2021pose, PIPCVPR2022, charles2016personalizing, BelagZ2016, kim2019pedx} is critical for applications in AR/VR, robotics, human-computer interaction, sports, and healthcare~\cite{hri_xu, wang2017realtime3dhumantracking}. Markerless motion capture methods typically use either cameras~\cite{wham:cvpr:2024,shen2024gvhmr,wang2025prompthmr,li2025genmo} or LiDAR~\cite{ren2024livehps, Ren2024LiveHPSRA, cai2023pointhps, jang2023movin}, but single-modality methods suffer from limited information: Monocular cameras lack depth, while LiDAR lacks texture.
Fusing both modalities provides complementary strengths, geometry and location information from LiDAR and appearance information from cameras
, enabling more precise human motion estimation of local pose and global translation.

Nevertheless, while multi-modal sensor methods provide richer information, they face increasing challenges as the number of sensors and modalities grows. On one hand, most multi-modal MoCap methods~\cite{motion2fusion, DeepVolumetric_2018ECCV, ma2021transfusioncrossviewfusiontransformer, xue2024freecaphybridcalibrationfreemotion} rely on data alignment via calibration, which propagates errors, especially when sensors have varying or dynamically changing perspectives, and thus requires fixed sensor placement with highly overlapping views, limiting flexibility and reducing the benefit of diverse viewpoints for occlusion handling. Although the most recent SoTA FreeCap~\cite{xue2024freecaphybridcalibrationfreemotion} eliminates manual calibration through on-the-fly estimation from matched keypoints, the involved calibration step can still accumulate errors, potentially leading to failures.
On the other hand, prior methods degrade significantly under severe noise~\cite{ren2024livehps}. Although recent methods such as LiveHPS++~\cite{Ren2024LiveHPSRA} leverage LiDAR geometry to improve translation estimation, severe noise can cause normalization of point clouds to shift human points away from the valid data domain, resulting in substantial inaccuracies and unstable trajectories.

The above limitations in prior works motivate two central research questions: 1) \textit{Can hybrid (LiDAR+Camera) motion capture achieve robust multi-modal alignment/matching under arbitrary sensor deployments?} Solving this challenge enables flexible deployment, where the system can adaptively adjust sensor layouts to mitigate occlusions and satisfy diverse task requirements in real-world applications; and 2) \textit{How to make precise prediction of global human translation in noisy and cluttered environments?}

To address these challenges, we propose \textbf{Sen-Cap}, a novel human motion capture approach that predicts (1) human local poses through a \textbf{Unified Across-Sensor Motion Estimator (\textbf{UAME})} and (2) global trajectories with a \textbf{Noise-Resilient Trajectory Tracker (\textbf{NTT})}, achieving robust and precise results even in noisy and variable-viewpoint environments. Here, the sensor-flexible multi-modal fusion strategy is instantiated within UAME rather than as an independent module. The local poses and global trajectories together constitute the final estimation of human motion.
Specifically,
\textbf{The Unified Across-Sensor Motion Estimator} estimates the human local pose and shape in the \textit{Human-Centric Space}. UAME consists of two coupled components, Human-Centric Space Alignment and Adaptive Sensor Fusion, which together form our sensor-flexible multi-modal fusion strategy. This design dynamically unifies data from a flexible number of uncalibrated sensors with unknown poses, enabling easy deployment in complex indoor and large-scale outdoor environments.
\textbf{The Noise-Resilient Trajectory Tracker} employs an iterative refinement mechanism that gradually expands the searching space, and even if the distribution of point clouds is greatly affected by extreme noise, it can achieve consistent and accurate global translation estimation.
They together result in a sensor-flexible and noise-resilient global pose estimation that supports variations in sensor modality, viewpoint and quantity without model retraining.

We evaluate \textit{Sen-Cap} against SoTA single-sensor and hybrid-sensor methods on the large-scale multi-person dataset Human-M3~\cite{fan2023human} and the multi-view dataset FreeMotion~\cite{ren2024livehps}. We further conduct cross-dataset evaluations on the challenging outdoor dataset LiDARHuman26M~\cite{li2022lidarcap} and the fast-motion dataset RELI11D~\cite{RELI11D}, where viewpoint discrepancies relative to the training data are substantial. Extensive experiments demonstrate that our approach achieves strong robustness to noise and generalization capability, while supporting a flexible number of sensors without model retraining.
Finally, we evaluate sensor flexibility by varying the number of sensors under a unified parameter setting, and assess noise resilience across different noise levels. The results confirm the adaptability of our framework to flexible deployment scenarios and its robustness to noise.
Our main contributions are summarized as follows:
\begin{enumerate}\setlength{\itemsep}{-2pt}
\item To the best of our knowledge, we present the first global human motion capture framework that supports a flexible number of multi-modal sensors (LiDAR and cameras) within a unified model.
\item We propose a Unified Across-Sensor Motion Estimator (including Human-Centric Space Alignment and Adaptive Sensor Fusion) and a Noise-Resilient Trajectory Tracker that enable cross-modal alignment with uncalibrated, multi-modal, varying numbers of sensors, while remaining robust to substantial sensor noise.
\item We validate performance of \textit{Sen-Cap} extensively on multiple public benchmarks as well as through real-world deployments, demonstrating unprecedented flexibility and robustness for in-the-wild applications.
\end{enumerate}

%% file: sec/2_related_work.tex
\section{Related Work}
\subsection{Visual Sensor-based Motion Capture}
Early motion capture methods that estimate high-quality human motions rely on wearable sensors, such as markers~\cite{optitrack,raskar2007prakash,loper2014mosh,Park2008,SongGodoy2016, UnstructureLan}, IMUs~\cite{ren2023lidar,yi2021transpose,huang2018DIP}, and ego cameras~\cite{EgoCapData2016,BetanDBMRR2016,JiangG2016,cao2017egocentric, Zheng2018HybridFusion} but are limited by issues such as sensor drift, discomfort for the wearer, and challenges in capturing certain movements, especially in dynamic or complex environments. The shift towards non-wearable motion capture has been significant with the use of cameras. Monocular-based methods~\cite{PonsMFR2014,HMR18,goel2023humans,mono-3dhp2017,alldieck2017optical,mehta2017monocular} provide convenient solutions but suffer from depth perception limitations and rely on single-modal representations that lack robustness. Multi-view-based methods~\cite{motion2fusion, DeepVolumetric_2018ECCV, malleson2019real} can capture comprehensive motion information but are difficult to set up and typically employ simple feature concatenation rather than principled cross-modal representation learning. Both of them are hard to recover the global human motions due to limited representation capacity. Recent efforts~\cite{wham:cvpr:2024,rajasegaran2022tracking,shen2024gvhmr,wang2024tram,TRACE,yuan2022glamr,ye2023slahmr,wang2025prompthmr,li2025genmo} that use dynamic monocular views to estimate global motions still face challenges with translation accuracy, particularly in long video sequences, primarily due to insufficient temporal representation modeling. LiDAR has been instrumental in capturing global human motion in open and large-scale environments~\cite{10472936, fan2023lidar}. Early LiDAR-based methods like LiDARCap~\cite{li2022lidarcap} focus on local pose estimation using point-based representations, while LiDAR-HMR~\cite{fan2023lidar} aims to reconstruct human meshes through geometric feature learning. However, these methods are limited by the lack of unified representation spaces that can effectively integrate geometric and semantic information. PointHPS~\cite{cai2023pointhps} and LiveHPS~\cite{ren2024livehps} have since advanced to predict full SMPL parameters, capitalizing on the depth information provided by LiDAR through improved point cloud. LiveHPS++~\cite{Ren2024LiveHPSRA} further improves the robustness and smoothness of human motion capture via temporal consistency modeling. Despite these advancements, challenges for LiDAR-based methods still remain due to data sparsity and the absence of texture information, fundamentally stemming from the lack of cross-modal representation learning capabilities.

\subsection{Hybrid Sensor-based Motion Capture}
Recognizing the limitations of single visual sensor-based approaches, research has increasingly focused on using multiple sensors~\cite{xu2023human}. However, existing multi-modal approaches typically employ naive feature concatenation or late fusion strategies, lacking principled cross-modal representation learning frameworks. Human-M3~\cite{fan2023human} integrates LiDARs with cameras to address occlusion challenges, and fully utilize advantages in different modal data through simple feature aggregation. However, these systems are often constrained by the need for precise calibration, necessitating fixed and pre-deployed setups, and more importantly, they lack unified cross-modal representation spaces that can effectively bridge the semantic gap between different modalities. A recent advance is FreeCap~\cite{xue2024freecaphybridcalibrationfreemotion}, which proposes a calibration-free method to match multi-persons in point clouds and images and uses the matching pairs to calculate a coarse calibration matrix. However, the network is heavyweight due to its lack of efficient representation learning mechanisms, and in cases where reference human points are few, the calibration matrix is inaccurate, degrading the performance of the fusion results. Additionally, it only supports one LiDAR within its framework, limiting its applicability and scalability. In contrast to these approaches that rely on explicit calibration or simple feature fusion, our method introduces a novel multi-modal representation learning framework that learns unified cross-modal representations through Human-Centric Space Alignment, eliminating the need for calibration estimation while supporting extendable LiDARs and cameras.

%% file: sec/3_method.tex
\begin{figure*}[t]
	\centering
	\includegraphics[width=\linewidth]{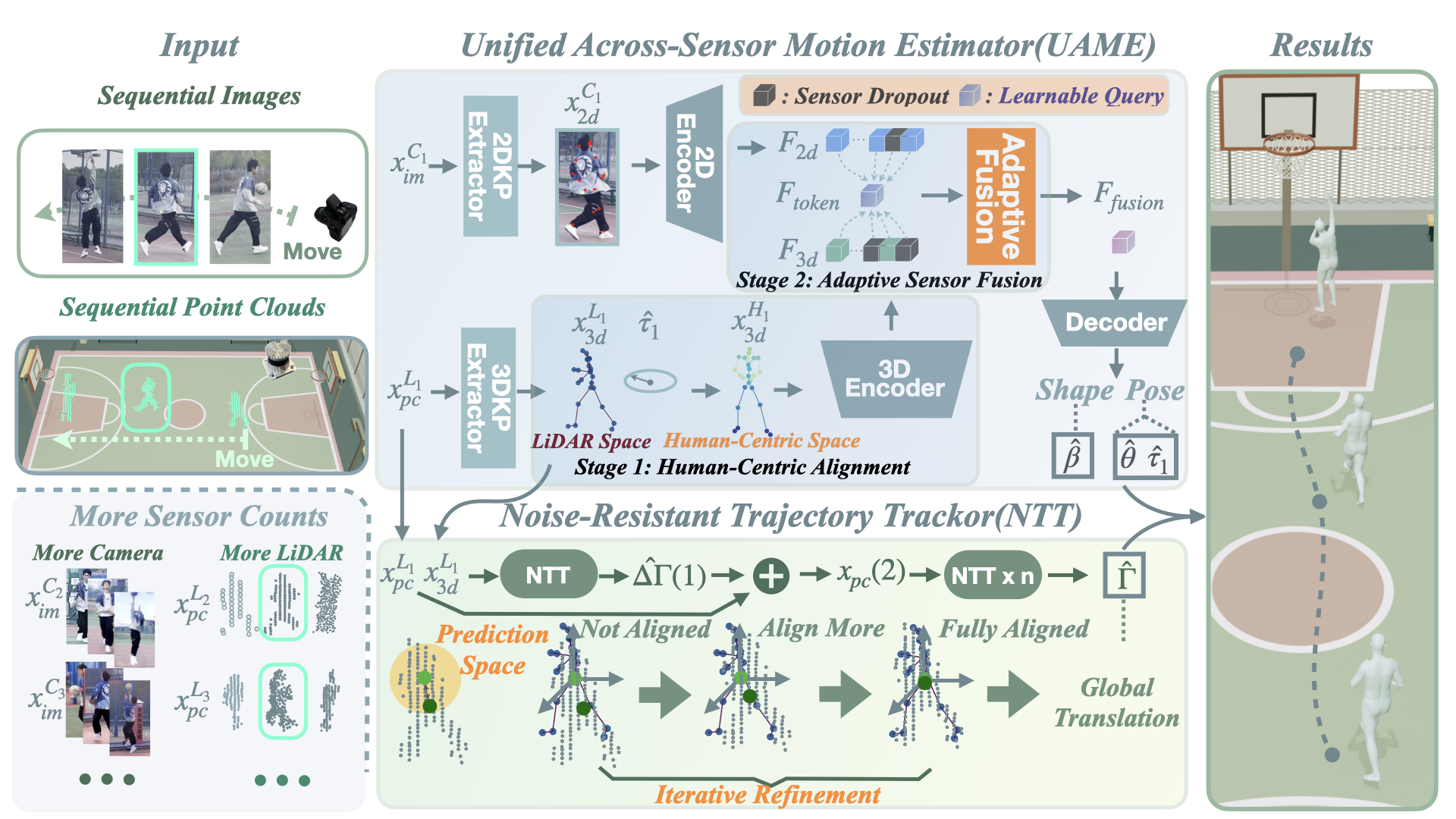}
	\caption{Overview of \text{Sen-Cap}. With sequential point clouds from movable LiDARs and 2D key points from movable cameras as input, Sen-Cap consists of two primary modules, a Unified Across-Sensor Motion Estimator (UAME) for obtaining the human global orientation, pose, and shape, and a Noise-resistant Trajectory Tracker (NTT) for global trajectory estimation.
    All components corresponding to our key contributions are highlighted in \textcolor{orange}{orange}.}
	\label{fig:pipeline}
\end{figure*}
\section{Methodology}
\label{sec:method}

Accurate 3D human motion capture from uncalibrated, heterogeneous sensors is challenging due to the inherent variability in sensor configurations and the susceptibility to noise in real-world environments. Existing methods often fail to address two critical issues: (1) \textit{dependency on fixed sensor setups} that requires precise calibration and cannot adapt to dynamic sensor availability; (2) \textit{sensitivity to sensor noise and occlusion} that leads to inaccurate global trajectory estimation.

\textbf{Sen-Cap} overcomes these limitations via two technical innovations: first, a \textbf{Unified Across-Sensor Motion Estimator}(Section~\ref{sec:local}) that extracts consistent human pose and shape from flexible sensor combinations in a calibration-free manner. In this work, ``calibration-free'' means that no explicit extrinsic parameters between sensors are required during inference. Our model does not assume pre-calibrated relative poses and can operate under arbitrary unknown rigid transformations between sensors; second, a \textbf{Noise-resistant Trajectory Tracker}(Section~\ref{sec:global}) that resists sensor noise through iterative refinement.
Notably, our method supports input from varying combinations of LiDARs and cameras, offering high flexibility and adaptability in real-world deployments.
An Overview of \textbf{Sen-Cap} is shown in Fig.~\ref{fig:pipeline}.

\paragraph{Problem Statement and Preliminaries} \quad
\textit{Sen-Cap} takes as input sequentially sampled point clouds ${x}_{\text{pc}}$ from LiDAR and normalized 2D keypoints ${x}_{\text{2d}}$ from images,  requiring no calibration and supporting a flexible number of sensors, and predicts the full SMPL~\cite{SMPL2015} parameters as the outputs, which represent the underlying human pose and shape.
Although our input is time-series data, for the sake of convenience, we save time $t$ when writing symbols. First, we follow LiveHPS++~\cite{Ren2024LiveHPSRA} to use the farthest point sampling algorithm (FPS) to sample the point cloud as a fixed number $x^{L_i}_{pc} \in \mathbb{R}^{256\times3}$ from the i-th LiDAR, and follow WHAM~\cite{wham:cvpr:2024} to normalize the 2D keypoints $x^{C_j}_{2d} \in \mathbb{R}^{17\times2}$ with its bounding box center and scale from the j-th camera. Meanwhile, following ~\cite{xue2024freecaphybridcalibrationfreemotion}, the ground truth of our algorithm is SMPL~\cite{SMPL2015} parameters with shape $\beta \in \mathbb{R}^{10}$, body pose $\theta \in \mathbb{R}^{23\times3}$, global orientation $\tau \in \mathbb{R}^{3}$ and global translation $\Gamma \in \mathbb{R}^3$ while $\hat{\beta}$, $\hat{\theta}$, $\hat{\tau}$ and $\hat{\Gamma}$ are corresponding predictions of our framework. These parameters are used to describe the mesh of the human body with 6890 vertices and 24 joints.
Finally, as cross-sensor human matching is essential for our sensor-flexible framework in multi-person scenarios, we build upon the optimization-based per-frame matching mechanism of FreeCap~\cite{xue2024freecaphybridcalibrationfreemotion}. To enable real-time online inference, we further introduce a matching memory bank. More details are provided in supplementary material.

\subsection{Unified Across-Sensor Motion Estimator: Local Pose Recovery}
\label{sec:local}
The core challenge in sensor-flexible motion capture is reconciling data from arbitrarily configured sensors. Our solution exploits the invariant properties of human kinematics: by estimating and aligning to the human centric space with \textit{Unified Across-Sensor Motion Estimator (UAME)}, we create a consistent reference frame that automatically unifies inputs from diverse sensors, regardless of their number, type, or viewpoint.
Architectural details of the Unified Across-Sensor Motion Estimator (UAME) are provided in supplementary material.

\textit{Limitation of prior local-motion methods.} \quad
Most prior multi-modal local pose pipelines(~\cite{optitrack}, ~\cite{xue2024freecaphybridcalibrationfreemotion}) either depend on explicit calibration to align sensor observations or adopt unconstrained feature aggregation under fixed sensor assumptions. As a result, they are sensitive to missing sensors and unstable cross-modal correspondences when the deployment changes at inference time.

As shown in Fig.~\ref{fig:pipeline}, UAME takes as input uncalibrated sequential 3D point clouds $x^{L_{i}}_{\text{pc}}$ from multiple LiDARs $L_{i}$, and 2D keypoints $x^{C_j}_{\text{2d}}$ from multiple cameras $C_j$. It then estimates the local pose $\hat{\boldsymbol{\theta}}$ and human shape $\hat{\boldsymbol{\beta}}$ in a human-centric coordinate system that remains consistent across diverse sensor viewpoints and modalities. This is achieved through two key stages: 1) \textit{Human-Centric Space Alignment}, and
2) \textit{Adaptive Sensor Fusion} addresses sensor variability through a bottleneck attention module that explicitly compresses multi-sensor information into a learnable latent token. By forcing all modality-specific features to interact through this shared bottleneck representation, the module adaptively re-weights sensor contributions according to their reliability and contextual relevance. The resulting fused feature dynamically adjusts to the current sensor configuration, enabling robust and accurate local pose estimation across diverse sensing conditions.

\paragraph{\textbf{Stage 1. Human-Centric Space Alignment} }
\quad
Given sequential point clouds $\{x_{pc}^{L_i}\}_{i=1}^{N_L}$ from $N_L$ LiDARs, we first learn a unified representation space that aligns all heterogeneous sensor modalities to a Human-Centric coordinate system $H$. This alignment mechanism enables effective cross-modal knowledge transfer without explicit calibration. The Human-Centric coordinate system is centered around humans and is independent of sensor placement. For each LiDAR $L_i$, a Gated Recurrent Unit (GRU) learns to predict rotation vector $\tau_i$ that transforms its 3D joint representations $x_{3d}^{L_i}$ encoded by 3D PointNet Encoder to the unified Human-Centric space $H_i$.
\begin{equation}
    \begin{aligned}
    x_{3d}^{H_{i}} = R(\hat{\tau_{i}})^{-1}x_{3d}^{L_{i}},
    \end{aligned}
\end{equation}
where $R(\cdot)$ transforms rotation vector to rotation matrix.
We then use loss $\mathcal{L}_{align}$ to optimize 3D representations in both LiDAR space $\mathbf{x}_{3d}^{L}= \{\mathbf{x}_{3d}^{L_1},\cdots,\mathbf{x}_{3d}^{L^{N_L}} \}$ and human-centric spaces $\mathbf{x}_{3d}^{H} = \{\mathbf{x}_{3d}^{H_1},\cdots,\mathbf{x}_{3d}^{H^{N_L}}\}$, together with 6D representation of the global transformation matrix~\cite{Zhou2018OnTC} $\tau_i^{6d} = \{\mathbf{\tau}_1^{6d},\cdots,\mathbf{\tau}_{N_L}^{6d} \}$:
\begin{equation}
    \begin{aligned}
    \mathcal{L}_{align}(\tau^{6d}, \mathbf{x}_{3d}^{L}, \mathbf{x}_{3d}^{H}) = & \frac{1}{N_L} \sum_{i=1}^{N_L} (\| \tau_i^{6d} - \hat{\tau_i}^{6d} \|^2_2
    +\| x_{3d}^{L_i} - \hat{x}_{3d}^{L_i} \|^2_2 + \| x_{3d}^{H_i} - \hat{x}_{3d}^{H_i} \|^2_2),
    \end{aligned}
\end{equation}
Besides, the 2D inputs are aligned by encoding the 2D keypoints $x_{2d}$ into a human-centric feature space via the 2D Encoder.

\paragraph{\noindent\textbf{Stage 2. Adaptive Sensor Fusion.}}
This stage introduces a bottleneck attention fusion module to explicitly mitigate sensor variability. Instead of directly performing unrestricted cross-modal interactions, modality-specific features are squeezed into a shared learnable latent token that serves as an information bottleneck. Through this constrained aggregation process, the model selectively preserves sensor-specific cues that are reliable and motion-relevant, while suppressing noisy or degraded signals. The resulting compact fused representation remains robust to changing sensor configurations, ensuring stable and accurate local pose estimation.

We first learn modality-specific representations that capture the unique characteristics of each sensor type.
The aligned 3D joint representations $x_{3d}^{H_i}$ are processed by a GRU to extract temporal motion features $F_{3d}$ that encode spatial-temporal dynamics.
Meanwhile, sequential 2D keypoints $\{x_{2d}^{C_j}\}_{j=1}^{N_C}$ from $N_C$ cameras are encoded by another GRU to produce visual motion features $F_{2d}$.

To mitigate sensor variability and enforce structured cross-modal interaction, we introduce a bottleneck attention-based adaptive fusion module.
Instead of allowing unrestricted pairwise fusion across modalities, we employ a learnable fusion token $F_{token}$ as a shared information bottleneck.
This token serves as the query in a cross-attention operation to aggregate multi-modal features, thereby squeezing heterogeneous sensor representations into a compact latent space:

\begin{equation}
F_{fusion} = \text{CrossAttn}(Q=F_{token},\, K=F_{m},\, V=F_{m}),
\end{equation}

where $F_{m} = \text{Concat}(F_{3d}, F_{2d})$ denotes the concatenated multi-modal features.
Through this bottleneck aggregation, the model adaptively reweights modality contributions according to their reliability and contextual relevance, while preventing direct and potentially noisy cross-modal interactions.
The resulting compact fused representation $F_{fusion}$ encodes the most informative sensor cues under the current configuration.

The fused features are then decoded by a GRU to regress SMPL parameters $\hat{\beta}, \hat{\theta}$:

\begin{equation}
\hat{\beta},\hat{\theta} = \text{GRU}(F_{fusion}),
\end{equation}

with the pose supervision defined as

\begin{equation}
\mathcal{L}_{pose}(\hat{\beta}, \hat{\theta})
= \alpha\|\beta-\hat{\beta}\|_2^2
+ \|\theta-\hat{\theta}\|_2^2.
\end{equation}

\textit{Sensor Dropout} \quad
Motivated by the above limitation, we introduce Sensor Dropout to explicitly train UAME under incomplete or degraded sensor configurations. Specifically, for each training sample, we randomly mask a certain percentage of feature tokens from both LiDAR and camera modalities before fusion. This training strategy simulates a spectrum of failures, from partial observation loss to full sensor absence, forcing the bottleneck fusion module to re-weight remaining reliable cues. The scaling-down and cross-configuration evaluations in Section~\ref{sec:robustness} are consistent with the intended robustness improvement without retraining.

\subsection{Noise-resistant Trajectory Tracker: Robust Global Localization}
\label{sec:global}
\begin{wrapfigure}{r}{0.50\linewidth}
    \centering
    \includegraphics[width=0.95\linewidth]{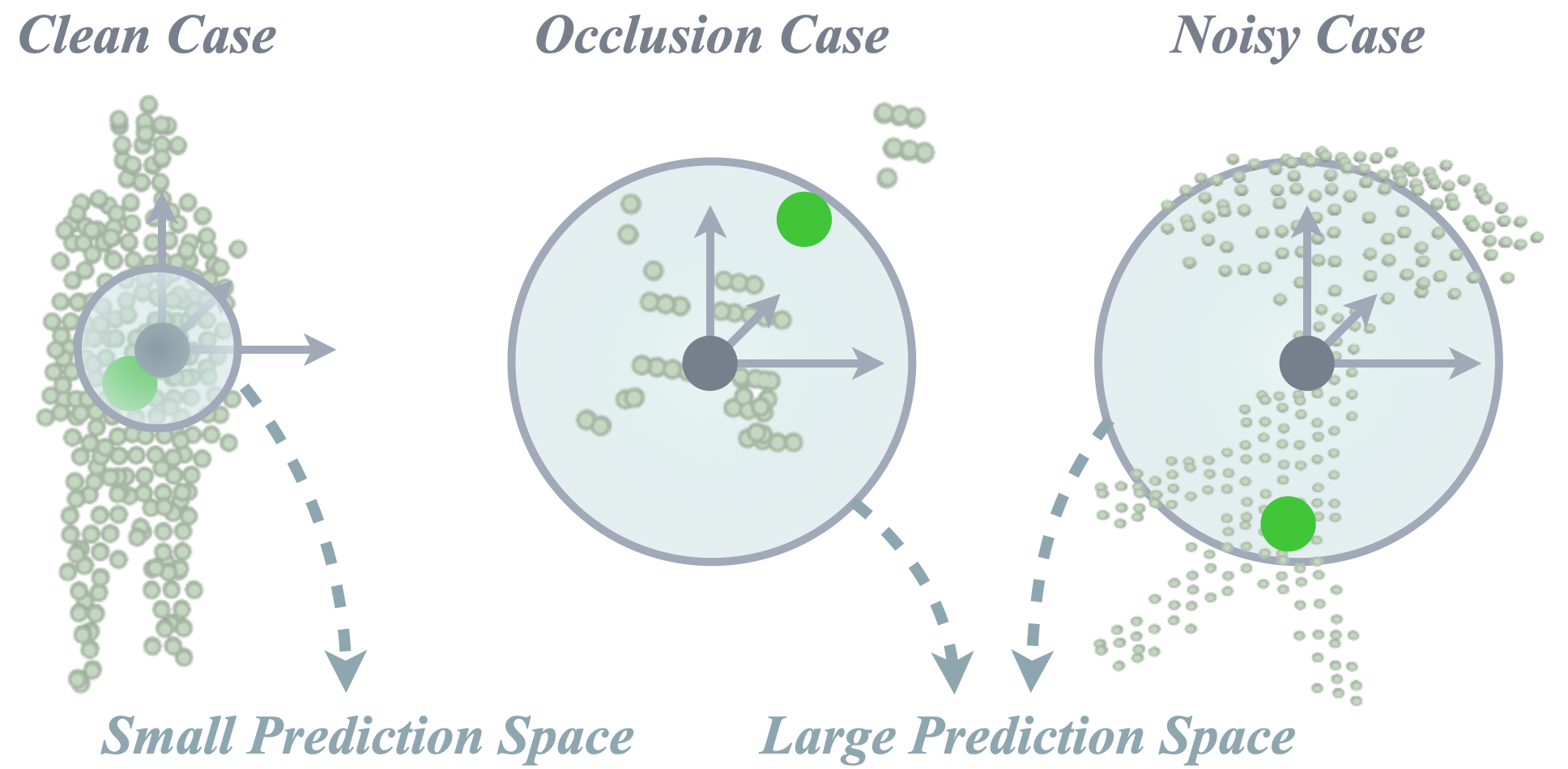}
    \caption{Systematic drift arises when the true root joint (green ball) is offset from the point cloud center $\overline{x}_{pc}$ (gray ball), especially under severe occlusion and noise. }
    \label{fig:transl}
\end{wrapfigure}
Besides a precise local motion,
an accurate global trajectory is another critical component for coherent human motion capture.
Thus, we then take as input the sequential point clouds $\mathbf{x}_{pc}$ from each LiDAR, and the corresponding 3D joints $\hat{\mathbf{x}}_{3d}^L$ (Section~\ref{sec:local}) to estimate the Global Trajectory $\hat{\Gamma}$.

\textit{Limitation of prior works.} \quad
Standard pipeline~\cite{ren2024livehps} first normalized the point cloud so that its centroid coincides with the coordinate origin, then predicts a single offset to reach the human root. This procedure implicitly assumes that the centroid is close to the pelvis; however, noise or partial occlusion can shift the centroid far away, forcing the network to explore an abnormally large \textbf{prediction space} of candidate offsets. The enlarged prediction region biases the learned prior toward under-estimating large displacements and finally produces the systematic drift visualized in Fig.~\ref{fig:transl}.

Hence, to keep the prediction space more stable under different perturbation levels, we adopt an iterative refinement strategy that gradually moves the centroid closer to the true root, reducing both offset magnitude and prediction variance at every step. Specifically, we use a Noise-resistant Trajectory Tracker (NTT) with an iterative refinement mechanism to obtain a robust and smooth global trajectory under sensor noise and environmental variations. For each step, we predict the offset $\Delta\hat{\Gamma}^n$ from the previous point cloud $x_{pc}^{n-1}$ at iteration $n-1$ and local motion $x_{3d}^L$, with

\noindent 1) \textit{Iterative Offset Prediction}: For each step $n$:
\begin{equation}
\Delta\hat{ \Gamma}^n = E_O(x_{pc}^{n-1}, x_{3d}^L), \quad x_{pc}^n = x_{pc}^{n-1} + \Delta \hat{\Gamma}^n,
\end{equation}
where $E_{O}$ is the offset predictor and we denote $x_{pc}^{0} = x_{pc}$ and the center of original point cloud $\overline{x}_{pc}$.

\noindent2)\textit{Constrained Optimization}: The final  $\hat{\Gamma}$ of the human in LiDAR coordinates is then determined as $\hat{\Gamma} = \overline{x}_{pc} + \sum_{n=1}^{N} \Delta \hat{\Gamma}^n$ after $N$ iterations.
We optimize the global trajectory representation with loss $\mathcal{L}_{tr}(\Gamma)$ using ground truth step offset and total trajectory, along with penalizing too large predictions,
\begin{equation}
\begin{aligned}
  \mathcal{L}_{tr}(\Gamma) &=  \| \hat{\Gamma} - \Gamma \|_2^2 + \lambda \frac{1}{N} \sum_{n=1}^N \mathcal{L}_{dtr}(\Delta \hat{\Gamma}^n),
\end{aligned}
\end{equation}
where $\mathcal{L}_{dtr}(\Delta \hat{\Gamma}^n)$ is the penalty loss for preventing large step size with penalty $\delta$:
\begin{equation}
  \mathcal{L}_{dtr}(\Delta \hat{\Gamma}^n) =
   \begin{cases}
   0  &, \left\| \Delta \hat{\Gamma}^n \right\| \leq \delta \\
   \left\| \left\| \Delta \hat{\Gamma}^n \right\| - \delta \right\|  &, \left\| \Delta \hat{\Gamma}^n \right\| > \delta.
   \end{cases}
\end{equation}

\textbf{Remarks.} \quad
Unlike the coordinate system-independent parameters ($\hat{{\beta}}, \hat{\theta}$) estimated in the local representation (Section~\ref{sec:local}),
the noise-resilient global trajectory $\hat{\boldsymbol{\Gamma}}$ is defined relative to LiDAR coordinate systems.
To this end, the Noise-resistant Trajectory Tracker(NTT) predicts the global translation for each LiDAR in a batched manner, depending on the number of available LiDARs.

\textbf{Real-Time Online Deployment.}\quad
For real-time online deployment, we additionally enhance temporal consistency and trajectory smoothness as detailed in supplementary material.

%% file: sec/4_Experiment.tex
\section{Experiments}
\label{sec:experiments}
\subsection{Experimental Setup}
\label{sec:exp_setup}

\paragraph{\noindent\textbf{Implementation Details.}}
Our framework is implemented in PyTorch and trained using AdamW with cosine learning rate decay.
All experiments are conducted on NVIDIA RTX-3090 GPUs.
Detailed hyperparameters and network architectures are provided in supplementary material.

\paragraph{\noindent\textbf{Baselines.}}
We compare against seven recent state-of-the-art methods.
Camera-based baselines are WHAM, GVHMR, GENMO, and PromptHMR~\cite{wham:cvpr:2024,shen2024gvhmr,li2025genmo,wang2025prompthmr}.
LiDAR-based baselines are LiveHPS and LiveHPS++~\cite{ren2024livehps,Ren2024LiveHPSRA}.
The camera--LiDAR hybrid baseline is FreeCap~\cite{xue2024freecaphybridcalibrationfreemotion}.

\paragraph{\noindent\textbf{Benchmarks and Protocols.}}
Following FreeCap~\cite{xue2024freecaphybridcalibrationfreemotion},
we conduct evaluations on the large-scale multi-person HumanM3~\cite{fan2023human} dataset and the multi-sensor FreeMotion~\cite{ren2024livehps} dataset.

\paragraph{\noindent\textbf{Metrics.}}
Following FreeCap~\cite{xue2024freecaphybridcalibrationfreemotion},
we report per-joint/vertex errors in millimeters (\textbf{J/V Err(PS/PST)}),
acceleration error in $m/s^2$ (\textbf{Accel Err}),
per-global-joint rotation error in degrees (\textbf{Ang Err}),
and scene-level unidirectional Chamfer distance in millimeters (\textbf{SUCD}).

\subsection{Quantitative Evaluations}
\label{sec:benchmark}

\begin{table}[t]
\centering
\resizebox{1.0\linewidth}{!}{
\setlength\tabcolsep{1pt}
\begin{tabular}{cc|cc|ccccc|ccccc}
\toprule
\multirow{2}{*}{\textbf{Methods}}& \multirow{2}{*}{\textbf{Venue\&Year}}& \multirow{2}{*}{\textbf{Modality}} & \multirow{2}{*}{\textbf{View}} & \multicolumn{5}{c|}{\textbf{Dataset FreeMotion}} & \multicolumn{5}{c}{\textbf{Dataset HumanM3~}} \\
\cmidrule(r){5-9} \cmidrule(l){10-14}
& &&& \textbf{J/V Err(PS)$\downarrow$} & \textbf{J/V Err(PST)$\downarrow$} & \textbf{Ang Err$\downarrow$} & \textbf{Accel Err$\downarrow$} & \textbf{SUCD$\downarrow$}
& \textbf{J/V Err(PS)$\downarrow$} & \textbf{J/V Err(PST)$\downarrow$} & \textbf{Ang Err$\downarrow$} & \textbf{Accel Err$\downarrow$} & \textbf{SUCD$\downarrow$} \\
\midrule
WHAM
~\cite{wham:cvpr:2024}
&CVPR'24 & C & T &
82.83/97.55 & - & 12.24 & 4.51 & - &
69.42/83.35 & - & 10.13 & \textbf{9.20} & -\\
GVHMR
~\cite{shen2024gvhmr}
&SA'24 & C & T &
72.47/84.23 & - & 13.80 & 4.59 & - &
79.39/91.23 & - & 13.49 & 19.00 & -
 \\
GENMO
~\cite{li2025genmo}
&ICCV'25 & C & T &
62.37/73.71 & - & 12.34 & 3.53 & - &
76.84/91.36 & - & 10.92 & 13.84 & -\\
PromptHMR
~\cite{wang2025prompthmr}
&CVPR'25 & C & T &
71.02/83.46 & - & 13.43 & 3.62 & - &
88.72/104.95 & - & 11.68 & 11.92 & -\\
LiveHPS
~\cite{ren2024livehps}
&CVPR'24 & L & T &
59.30/73.12 & 100.81/109.09 & 13.10 & 6.18 & 4.97&
57.81/71.27 & 97.11/103.10 & 10.44 & 12.58 & 6.75 \\
LiveHPS++
~\cite{Ren2024LiveHPSRA}
&ECCV'24 & L & T &
54.15/57.74 & 88.91/97.53 & 11.85 & 4.37 & 3.47 &
55.67/68.73 & 89.06/95.85 & 10.26 & 11.51 & \textbf{5.60}\\
FreeCap
~\cite{xue2024freecaphybridcalibrationfreemotion}
&AAAI'25 & L+C & T &

53.31/65.50 & 95.97/102.91 & 11.14 & 5.97 & 4.82 &
55.45/68.52 & 96.47/102.67 & 9.14 & 9.60 & 6.66 \\
\midrule

\textbf{Ours}$^\dagger$ &
- & L & T &
52.98/56.41 & 84.37/91.62 & 11.21 & 4.28 & 3.21 &
53.94/66.82 & 86.75/93.14 & 9.88 & 11.08 & 5.42\\

\textbf{Ours} &
- & L+C & T &
\textbf{47.46/57.84} & \textbf{75.16/81.65} & \textbf{9.94} & \textbf{3.01} & 3.83&
\textbf{42.59/52.09} & \textbf{73.50/77.53} & \textbf{5.71} & 10.18 & 5.74
 \\
\midrule
WHAM
~\cite{wham:cvpr:2024}
&CVPR'24 & C & N &
108.16/124.52 & - & 19.89 & 4.49 & - &
85.50/101.44 & - & 10.42 & \textbf{9.25} & - \\
GVHMR
~\cite{shen2024gvhmr}
&SA'24  & C & N &

114.00/126.41 & - & 18.22 & 4.32 & - &
83.02/95.83 & - & 14.68 & 20.14 & - \\
GENMO
~\cite{li2025genmo}
&ICCV'25 & C & N &
72.48/81.96 & - & 12.87 & 3.84 & - &
79.31/94.82 & - & 11.15 & 14.27 & -\\
PromptHMR
~\cite{wang2025prompthmr}
&CVPR'25 & C & N &
96.00/107.38 & - & 16.54 & 3.69 & - &
91.46/108.73 & - & 12.06 & 12.34 & -\\
FreeCap
~\cite{xue2024freecaphybridcalibrationfreemotion}
&AAAI'25 & L+C & N &
57.37/69.61 & 99.20/106.15 & 11.74 & 6.08 & 4.96&
56.24/69.25 & 96.45/102.46 & 9.17 & 9.59 & 6.58 \\
\midrule
\textbf{Ours} &
- & L+C & N &
\textbf{52.25/63.05} & \textbf{79.05/85.82} & \textbf{10.80} & \textbf{3.16} & \textbf{3.93} &
\textbf{42.64/52.28} & \textbf{73.41/77.51} & \textbf{5.76} & 10.15 & \textbf{5.74}\\
\bottomrule
\end{tabular}}
\caption{
Evaluation on large-scale FreeMotion and Human-M3 datasets.
The ``View'' specification differentiates between performance under training (T) and novel camera viewpoints (N).
In the ``Modality'' column, C denotes Camera, L denotes LiDAR, and L+C denotes the multi-modal setting.
Entries marked with ``-'' correspond to metrics that are inapplicable to camera-based methods due to inherent scale ambiguity.
\textbf{Ours}$^\dagger$ denotes our method using only a single LiDAR sensor.
Note that \textbf{Ours}$^\dagger$ achieves performance comparable to LiDAR-only methods (LiveHPS~\cite{ren2024livehps}, LiveHPS++~\cite{Ren2024LiveHPSRA}), while the full multi-modal version yields substantial gains, highlighting the effectiveness of our calibration-free architecture in aligning and fusing heterogeneous sensors.
}
\label{tab:compare_offline}
\end{table}
As summarized in Table~\ref{tab:compare_offline}, we report results under two evaluation settings:
(i) \textit{training view} and (ii) \textit{novel camera viewpoints}.
The training-view setting reflects standard in-distribution accuracy.
The novel-view setting, by contrast, introduces unseen viewpoints that effectively alter camera extrinsics at test time.
Evaluating under this protocol is particularly relevant to our formulation, as Sen-Cap does not rely on explicit calibration and is designed to remain stable under such viewpoint changes.

Under the training-view protocol, our Unified Across-Sensor Motion Estimator (UAME; Section~\ref{sec:local}) effectively fuses complementary cues from LiDAR and cameras,
leading to substantial improvements over camera-only (WHAM~\cite{wham:cvpr:2024}, GVHMR~\cite{shen2024gvhmr}, GENMO~\cite{li2025genmo}, PromptHMR~\cite{wang2025prompthmr}) and LiDAR-only (LiveHPS~\cite{ren2024livehps}, LiveHPS++~\cite{Ren2024LiveHPSRA}) baselines on most metrics.
Moreover, we outperform the hybrid method FreeCap~\cite{xue2024freecaphybridcalibrationfreemotion}, largely due to our human-centric alignment strategy that removes the need for explicit calibration and reduces calibration-induced errors.

We also report a LiDAR-only variant (\textbf{Ours}$^\dagger$) using a single LiDAR sensor.
It achieves performance comparable to LiDAR-only methods, while the full multi-modal version yields significant gains,
highlighting that Sen-Cap can effectively align and benefit from heterogeneous sensors in a calibration-free manner.

\subsection{Out-of-Domain Generalization}
\label{sec:cross_domain}
\begin{table}[t]
\centering
\resizebox{\linewidth}{!}{
\setlength\tabcolsep{3pt}
\begin{tabular}{c|ccccc|ccccc}
\toprule
\multirow{2}{*}{\textbf{Methods}} & \multicolumn{5}{c|}{\textbf{Dataset LiDARHuman26M}}&\multicolumn{5}{c}{\textbf{Dataset RELI11D}} \\
\cmidrule(r){2-6}
\cmidrule(r){7-11}
& \textbf{J/V Err(PS)$\downarrow$} & \textbf{J/V Err(PST)$\downarrow$} & \textbf{Ang Err$\downarrow$} & \textbf{Accel Err$\downarrow$} & \textbf{SUCD$\downarrow$}
& \textbf{J/V Err(PS)$\downarrow$} & \textbf{J/V Err(PST)$\downarrow$} & \textbf{Ang Err$\downarrow$} & \textbf{Accel Err$\downarrow$} & \textbf{SUCD$\downarrow$}\\

\midrule
WHAM~\cite{wham:cvpr:2024}
&
151.18/191.73&-&32.92&3.16&-
&106.13/126.71&-&22.10&2.70&- \\
GVHMR~\cite{shen2024gvhmr}
&
187.01/226.19&-&28.58&\text{3.04}&-&
111.73/131.73&-&21.37&2.42 &-
\\

LiveHPS~\cite{ren2024livehps}
&
160.70/199.68&224.17/247.18&29.15&9.94&8.45
& 84.29/104.24&100.90/113.20&24.03&3.47&5.48 \\

LiveHPS++~\cite{Ren2024LiveHPSRA}
&
141.03/179.68&196.30/221.71&26.23&4.67&5.20
&78.45/95.41&88.10/102.43&22.28&1.68&2.09 \\

FreeCap~\cite{xue2024freecaphybridcalibrationfreemotion}
&
160.87/202.42&224.35/249.65&28.53&10.16&9.65
&80.49/98.47&98.88/109.36&23.60&3.48&5.53 \\

\midrule
\textbf{Ours}&
\textbf{131.32/164.39}&\textbf{191.59}/\textbf{216.73}&\textbf{24.04}&\textbf{3.03}&\textbf{5.11}
&\textbf{61.23/73.28}&\textbf{81.11/85.48}&\textbf{17.87}&\textbf{1.39}&\textbf{1.48} \\
\bottomrule
\end{tabular}}
\caption{Cross-Domain Evaluation on Noisy and Fast-motion Benchmarks: the outdoor noisy LiDARHuman26M dataset and the indoor fast-motion RELI11D dataset. Our approach is trained on AMASS and FreeMotion.}
\label{tab:cross_dataset}
\end{table}

To evaluate out-of-domain generalization, we conduct cross-dataset experiments on
LiDARHuman26M~\cite{li2022lidarcap} and RELI11D~\cite{RELI11D}.
Our model is trained only on AMASS~\cite{AMASS_ICCV2019} and FreeMotion~\cite{ren2024livehps} without any fine-tuning on these datasets.

As shown in Table~\ref{tab:cross_dataset}, Sen-Cap achieves competitive or superior performance
across most metrics despite the domain gap.
These results indicate that our framework generalizes well to unseen environments.

\subsection{Qualitative Evaluations}
\label{sec:qualitative}

\begin{figure}[t]
	\centering
    \includegraphics[width=\linewidth]{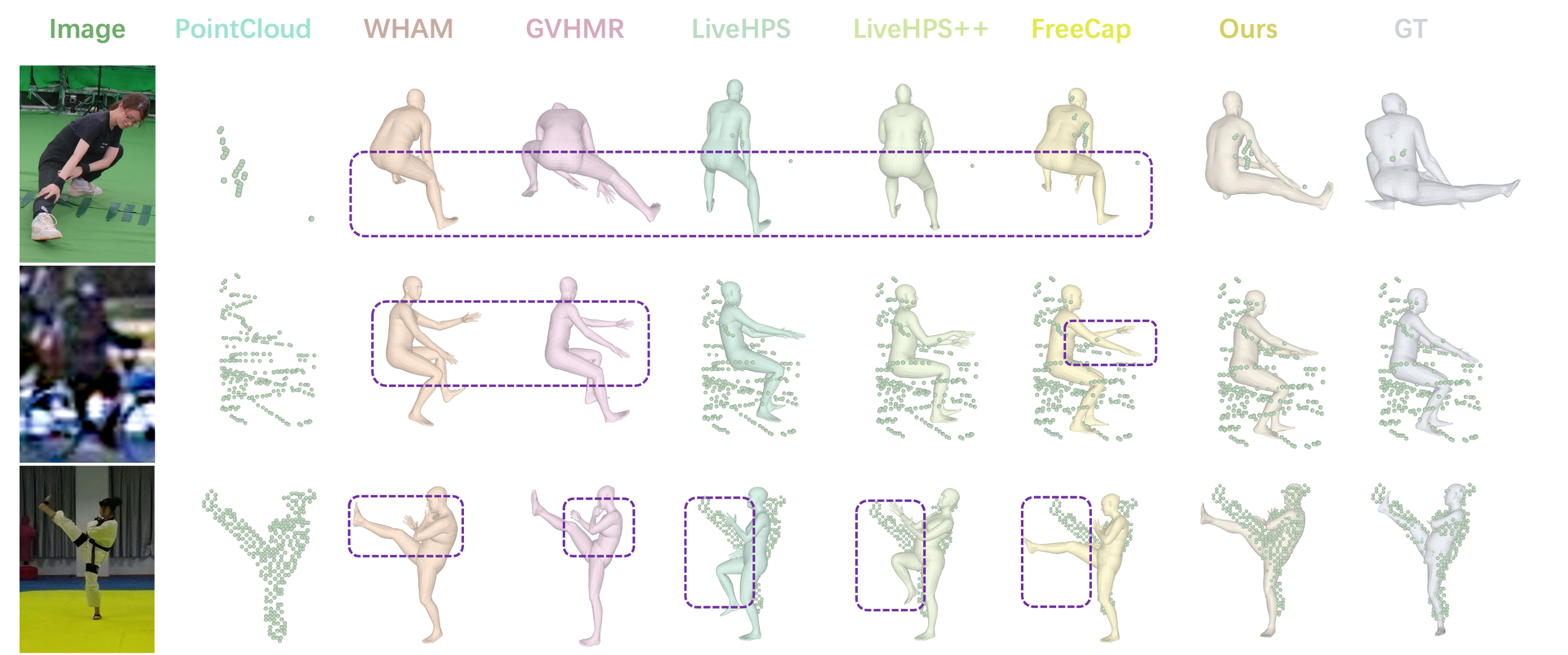}
    \caption{Qualitative comparisons showing \text{Sen-Cap}'s superior motion capture in challenging scenarios: (1) sparse-point leg stretching (2) Outdoor cycling (3) Fast-motion Taekwondo. We have highlighted the regions with significant differences using purple boxes.}
	\label{fig:compare_vis}
\end{figure}
Fig.~\ref{fig:compare_vis} provides qualitative comparisons in challenging scenarios, including
(1) sparse-point leg stretching, (2) outdoor cycling, and (3) fast-motion Taekwondo.
When LiDAR point clouds become sparse or partially missing, image priors help maintain plausible body shape.
Conversely, in outdoor conditions with degraded image quality, LiDAR geometry provides stable cues.
More qualitative results and trajectory analyses are included in supplementary material.

\subsection{Component Analysis}
\label{sec:ablation}

We conduct controlled ablations to isolate the contribution of each design choice in Sen-Cap.

\paragraph{\noindent\textbf{Ablations on UAME.}}
\begin{table}[t]
\centering
\setlength\tabcolsep{3pt}
\resizebox{\linewidth}{!}{
\begin{tabular}{c|c|ccccc}
\toprule
\multicolumn{1}{c}{} & \multicolumn{1}{c|}{} & \multicolumn{5}{c}{\textbf{Dataset FreeMotion}} \\
\cmidrule(r){3-7}
\multicolumn{2}{c|}{}
& \textbf{J/V Err(PS)$\downarrow$}
& \textbf{J/V Err(PST)$\downarrow$}
& \textbf{Ang Err$\downarrow$}
& \textbf{Accel Err$\downarrow$}
& \textbf{SUCD$\downarrow$} \\
\midrule

\multirow{2}{*}{\textbf{\makecell{Feature Space \\ (UAME)}}}
& LiDAR-Centric
& 68.73/81.42 & 92.38/100.81 & 12.75 & 3.39 & 5.11 \\
& Human-Centric
& \textbf{52.25/63.05} & \textbf{78.87/85.66} & \textbf{10.80} & \textbf{3.12} & \textbf{3.91} \\

\midrule
\multirow{3}{*}{\textbf{\makecell{Feature Fusion \\ (UAME)}}}
& Learnable Linear Fusion
& 60.63/72.41 & 85.49/93.09 & 11.73 & 3.27 & 4.45 \\
& Fixed-Weight Fusion
& 61.37/73.13 & 85.97/93.55 & 11.70 & 3.28 & 4.53 \\
& Bottleneck Attention (Ours)
& \textbf{52.25/63.05} & \textbf{78.87/85.66} & \textbf{10.80} & \textbf{3.12} & \textbf{3.91} \\

\midrule
\multirow{5}{*}{\textbf{\makecell{Trajectory \\ Estimation \\ (NTT)}}}
& repeat $\times$ 1
& 52.25/63.05 & 87.88/94.01 & 10.80 & 5.05 & 4.26 \\
& repeat $\times$ 2
& 52.25/63.05 & 82.42/88.78 & 10.80 & 4.18 & 3.54 \\
& repeat $\times$ 3
& 52.25/63.05 & 78.87/85.66 & 10.80 & 3.12 & 3.91 \\
& repeat $\times$ 4
& 52.25/63.05 & 74.88/82.20 & 10.80 & 3.17 & 3.81 \\
& repeat $\times$ 5
& 52.25/63.05 & 74.63/82.00 & 10.80 & 3.17 & 3.75 \\

\bottomrule
\end{tabular}
}
\caption{
Ablation studies on our design choices.
We analyze the impact of feature space, fusion strategy, and trajectory estimation iterations.
\textbf{Note that the local pose metrics (J/V Err(PS) and Ang Err) remain unchanged across different NTT settings},
since NTT refines global translation while keeping the estimated local pose fixed.
}
\label{tab:ab}
\end{table}
Table~\ref{tab:ab} analyzes key components in UAME.
First, the feature space choice is crucial: replacing our human-centric alignment with a LiDAR-centric feature space significantly degrades performance,
confirming the importance of human-centric representation for calibration-free operation.
Second, we evaluate alternative fusion schemes including a learnable linear aggregation (via concatenation and projection) and a static summation-based fusion.
Unlike these fully connected or fixed aggregation strategies, our bottleneck attention explicitly restricts cross-modal communication to pass through a compact latent token.
This structured information compression enables reliability-aware modality weighting and leads to improved robustness.

\begin{figure}[t]
	\centering
    \includegraphics[width=\linewidth]{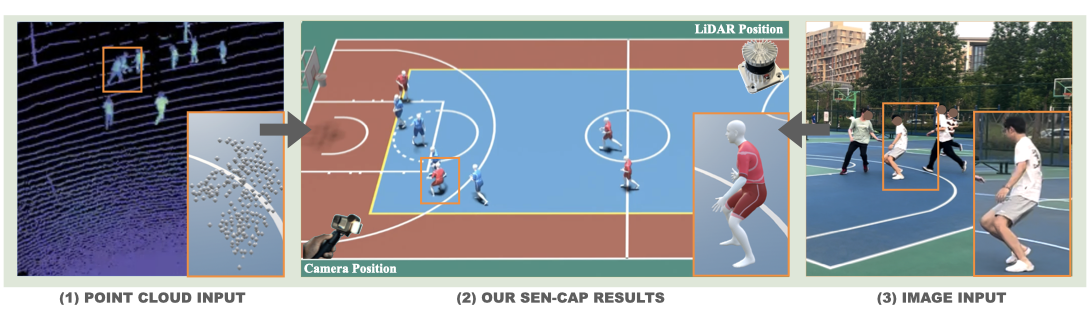}
	\caption{Application on a basketball example to illustrate the versatility of Sen-Cap. The results (2) confirm that Sen-Cap successfully fuses heterogeneous inputs from a camera (3) and a LiDAR sensor (1), enabling accurate 3D human pose estimation in challenging dynamic scenarios without requiring sensor calibration.}
	\label{fig:application}
\end{figure}

\paragraph{\noindent\textbf{Ablations on NTT.}}
We further evaluate the iterative refinement strategy in NTT.
As shown in Table~\ref{tab:ab}, the trajectory error progressively decreases as the number of refinement iterations increases,
with three iterations providing a favorable balance between accuracy and computational cost.
Note that local pose metrics (e.g., J/V Err(PS) and Ang Err) remain unchanged across different NTT settings,
since this module primarily refines global translation while keeping local pose estimation fixed.
Additional ablations and qualitative studies are provided in supplementary material.

\subsection{Robustness Evaluation under Sensor and Noise Variations}
\label{sec:robustness}

A core goal of Sen-Cap is to remain reliable under varying sensor configurations and noisy environments.
We validate this robustness through two targeted evaluations.

\subsubsection{Sensor Flexibility.}
\label{subsec:sensor_scalability}

\begin{figure*}[t]
    \centering
	\includegraphics[width=\linewidth]{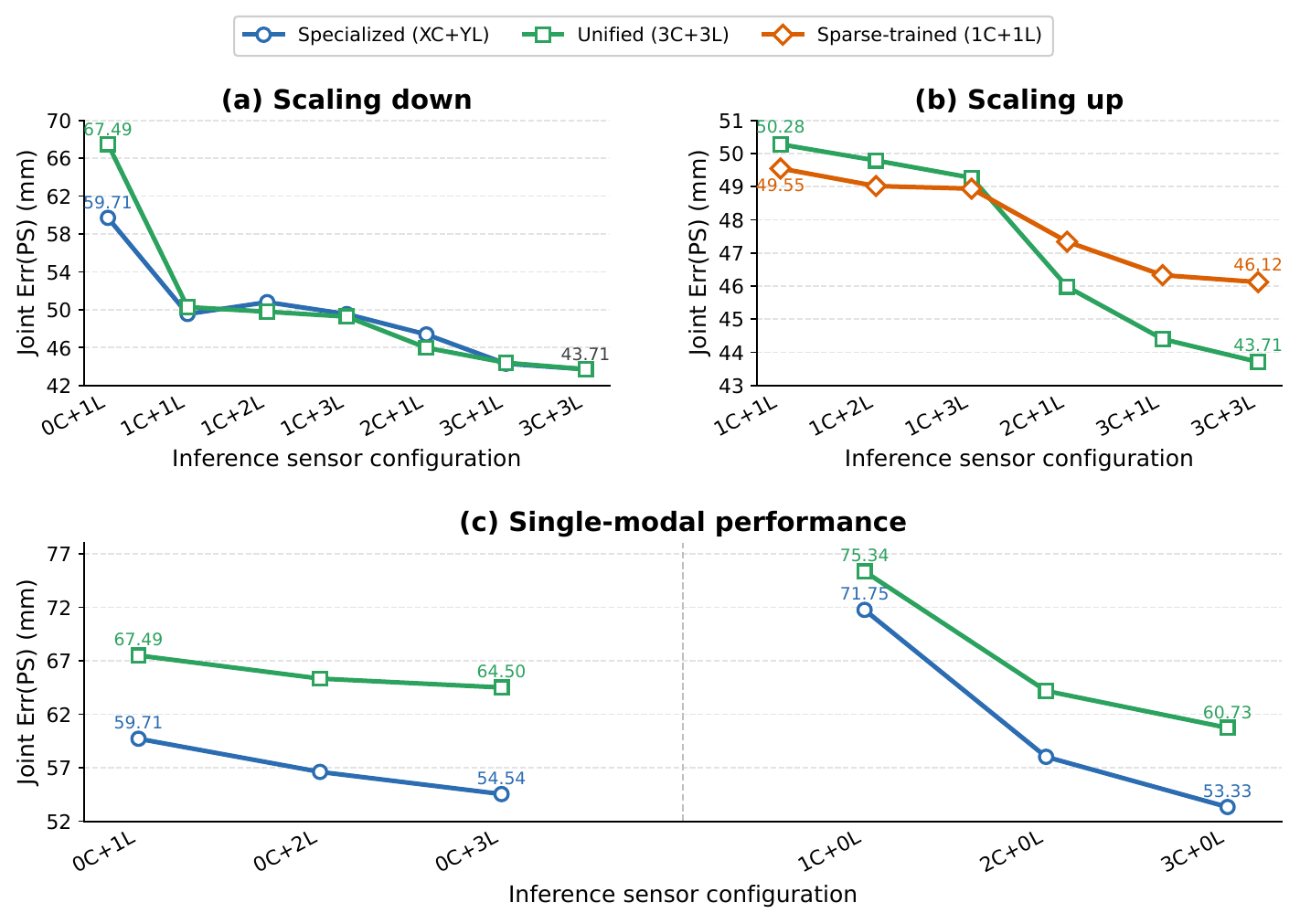}
	\caption{
	Sensor flexibility on FreeMotion-indoor.
	We visualize Joint Err(PS) for (a) scaling down from a unified 3C+3L model, (b) scaling up a sparse 1C+1L model with additional test-time sensors, and (c) single-modal LiDAR-only and camera-only configurations.
	Lower values are better.
	C and L denote the numbers of cameras and LiDARs.
	}
    \label{fig:sensor_flexibility}
\end{figure*}

Sen-Cap demonstrates strong sensor flexibility, which can adapt to varying numbers and modalities of sensors without retraining or recalibration. 
We evaluate this property from three complementary settings, including scaling down, scaling up, and single-modal inference. 
For clarity, Fig.~\ref{fig:sensor_flexibility} visualizes Joint Err(PS) under different inference sensor configurations.

\textbf{Scaling down.}
We evaluate a single unified model trained with multiple sensors (3 LiDARs + 3 cameras) under degraded test-time configurations with fewer sensors.
As shown in Fig.~\ref{fig:sensor_flexibility}(a), the unified model can effectively adapt to various reduced sensor configurations at test time.
Its performance remains comparable to models trained specifically under the corresponding configurations, demonstrating strong flexibility without retraining or recalibration.
This trend is consistent with the Sensor Dropout design in Section~\ref{sec:local}, which explicitly trains UAME to handle missing-sensor cases.

\textbf{Scaling up.}
We also train on a minimal sensor setup and evaluate with additional sensors at inference time.
As shown in Fig.~\ref{fig:sensor_flexibility}(b), adding cameras and/or LiDARs generally reduces the reconstruction error, indicating that Sen-Cap can leverage extra sensors at test time and generalizes gracefully to denser deployments.
We also include the unified 3C+3L model under the same inference configurations for comparison.

\textbf{Single-modal performance.}
Fig.~\ref{fig:sensor_flexibility}(c) shows that Sen-Cap maintains reasonable performance with either LiDAR-only or camera-only inputs.
For both LiDAR-only and camera-only settings, increasing the number of available sensors leads to better performance, showing that the framework can benefit from additional sensors even within a single modality.

Together, these results demonstrate that Sen-Cap is both robust and adaptable to diverse sensor configurations, supporting calibration-free deployment in real-world multi-sensor motion capture scenarios.

\subsubsection{Noise Resilience}
\label{subsec:noise_resilience}

To systematically evaluate robustness against clutter and object noise,
we construct three object-noise subsets of FreeMotion by adding synthetic clutter around the human point clouds:
(i) Level 1 contains low-noise samples;
(ii) Level 2 adds small objects (0.1--0.5\,m);
(iii) Level 3 adds larger objects (0.5--1\,m) to create moderate clutter.
The clutter is generated by sampling ShapeNet objects, scaling them to the target size, simulating visible points via raycasting from the LiDAR viewpoint,
and randomly placing the resulting object point clouds around the human.

\noindent
\begin{minipage}[t]{0.48\linewidth}
\centering
\setlength\tabcolsep{3pt}
\renewcommand{\arraystretch}{1.08}
\resizebox{\linewidth}{!}{
\begin{tabular}[t]{cc|ccc}
\toprule
\textbf{Noise} & \textbf{Setting}
& \textbf{\makecell{J/V Err\\(PST)$\downarrow$}}
& \textbf{Accel$\downarrow$}
& \textbf{SUCD$\downarrow$} \\
\midrule

\multirow{2}{*}{L1}
& w/o NTT & 67.91/72.18 & 3.64 & 3.41 \\
& w NTT  & \textbf{64.28/69.12} & \textbf{2.93} & \textbf{3.19} \\

\midrule

\multirow{2}{*}{L2}
& w/o NTT & 96.49/101.84 & 10.54 & 189.51 \\
& w NTT  & \textbf{89.49/95.33} & \textbf{3.08} & \textbf{185.45} \\

\midrule

\multirow{2}{*}{L3}
& w/o NTT & 367.51/374.16 & 94.62 & 553.86 \\
& w NTT  & \textbf{158.53/165.00} & \textbf{6.71} & \textbf{377.83} \\

\bottomrule
\end{tabular}
}

\captionsetup{type=table,hypcap=false,skip=2pt}
\captionof{table}{
Ablation of NTT under increasing noise levels.
}
\label{tab:ntt_noise_ablation}
\end{minipage}
\hfill
\begin{minipage}[t]{0.48\linewidth}
\raggedright
We compare performance with and without NTT ($N{=}3$).
As summarized in Table~\ref{tab:ntt_noise_ablation}, NTT consistently reduces errors across all noise levels,
with the largest gains under the most challenging clutter setting (Level 3).
Details of the noise-generation procedure and the definitions of each noise level are provided in the supplementary material.
\end{minipage}

\subsection{Real-world Deployment in Large-scale Scenes}
\label{sec:deployment}

In real-world large-scale motion capture, multi-sensor setups are often necessary, while flexible sensor placements make calibration difficult and error-prone.
Sen-Cap supports calibration-free deployment with varying sensor numbers and placements, enabling reliable motion capture in large scenes.
Fig.~\ref{fig:application} presents a real-world basketball example where Sen-Cap fuses LiDAR and camera observations to reconstruct accurate 3D human poses.
Further technical details and demonstrations are provided in supplementary material and the supplemental video.

%% file: sec/6_Conclusion.tex
\vspace{-2ex}
\section{Conclusions}
To conclude, we present \textit{Sen-Cap}, a motion capture framework that operates robustly with flexible combinations of LiDARs and cameras under noisy environments. It eliminates calibration through a Unified Across-Sensor Motion Estimator that aligns data in a human-centric space, and ensures noise robustness via a Noise-Resistant Trajectory Tracker with iterative refinement. Extensive experiments across multiple datasets and in the real world demonstrate state-of-the-art performance, validating the practicality and flexibility of our approach for real-world motion capture.